\documentclass[runningheads]{llncs}
\usepackage{graphicx}
\usepackage[utf8]{inputenc}
\usepackage{lineno,hyperref}
\modulolinenumbers[5]
\usepackage{amssymb}
\usepackage{graphicx}
\usepackage{dcolumn}
\usepackage{bm}
\usepackage{epstopdf}
\usepackage{amsmath}
\usepackage{textcomp} 
\usepackage{multirow}

\usepackage{times}  
\usepackage{helvet} 
\usepackage{courier}  
\usepackage[misc]{ifsym}
\usepackage{hyperref}
\newcolumntype{P}[1]{>{\centering\arraybackslash}p{#1}}
\begin{document}
\title{Energy consumption forecasting using a stacked nonparametric Bayesian approach}
%
%
%
 \author{Dilusha Weeraddana\inst{1}(\Letter)\and
Nguyen Lu Dang Khoa\inst{1}\and
Lachlan O'Neil\inst{2} \and
 Weihong Wang\inst{1}\and
 Chen Cai \inst{1}}


\authorrunning{D. Weeraddana et al.}
%

\tocauthor{Dilusha Weeraddana,
Nguyen Lu Dang Khoa,
Lachlan O'Neil,
 Weihong Wang,
 Chen Cai}

\institute{\textsuperscript{1}Data61-The Commonwealth Scientific and Industrial Research Organisation (CSIRO),
  Australia\\
  \textsuperscript{2}Energy-The Commonwealth Scientific and Industrial Research Organisation,
  Australia}
\maketitle              
\begin{abstract}
In this paper, the process of forecasting household energy consumption is studied within the framework of the nonparametric Gaussian Process (GP), using multiple short time series data.  As we begin to use smart meter data to paint a clearer picture of residential electricity use, it becomes increasingly apparent that we must also construct a detailed picture and understanding of consumer's complex relationship with gas consumption. Both electricity and gas consumption patterns are highly dependent on various factors, and the intricate interplay of these factors is sophisticated. Moreover, since typical gas consumption data is low granularity with very few time points, naive application of conventional time-series forecasting techniques can lead to severe over-fitting. Given these considerations, we construct a stacked GP method where the predictive posteriors of each GP applied to each task are used in the prior and likelihood of the next level GP. We apply our model to a real-world dataset to forecast energy consumption in Australian households across several states. We compare intuitively appealing results against other commonly used machine learning techniques. Overall, the results indicate that the proposed stacked GP model outperforms other forecasting techniques that we tested, especially when we have a multiple short time-series instances.

\end{abstract}
\section{Introduction}
\subsection{Energy forecasting across major Australian states}

Accurately forecasting residential energy usage is critically important for informing network infrastructure spending, market decision making and policy development. With concerns not only over the total volume of consumption within a time period, but also the maximum \lq peak\rq \ simultaneous consumption, this becomes a complicated topic to explore, one which the Australian Energy Market Operator (AEMO) explores yearly in their Electricity Statement of Opportunities (ESOO) \cite{aemoreport19} and Gas Statement of Opportunities (GSOO) \cite{aemoreport19gas}. Forecasting energy consumption with high accuracy is a challenging task. The energy consumption time series is non-linear and non-stationary, with daily and weekly cycles. In addition, it is also noisy due to the large loads with unknown hours of operation, special events and holidays, extreme weather conditions and sudden weather changes.

Energy forecasting has been an active area of research. There are two main groups of approaches: statistical, such Autoregressive Integrated Moving Average  (ARIMA),  exponential smoothing \cite{de2018forecasting}  and Machine Learning, with Tree based models, Neural Networks and Bayesian methods being the most popular representative of this group \cite{cheng2012random,pole2018applied}.

In Australia, interval meters typically record electricity usage in half hour intervals. This provides a rich source of data to develop forecasting models. However, the majority of electricity meters in Australia are basic accumulation meters which simply accumulate the amount of energy consumed over a longer timeframe, typically on a quarterly basis \cite{smartenergy,renew}. When it comes to gas metering, almost all meters in Australia are basic meters. This low penetration rate of interval meters in the electricity and gas sectors hinders the industry’s ability to forecast accurately.

Against this backdrop, this current study aims to develop a method which can be used to successfully predict residential energy consumption for individual households using readings from basic meters (and interval meters) and other available information such as weather or demographics.

Current research has calculated the average 10 year annual energy consumption growth for Australia to be 0.7\% for the residential sector, up until the 2016-17 financial year (Table 2.3 in \cite{AustralianEnergyUpdate2018}).  However, these estimates only consider aggregated load. There is still a poor understanding of changes in individual household energy consumption. This makes it difficult to build robust models which can capture unusual aggregate changes created by the chaotic nature of a system made up of millions of households. As such, the key objective of this work is to collaborate with leading Energy utilities and make use of machine learning techniques to identify future gas and electricity consumption for individual households. In this research, more than 2,500 households are studied across different states of Australia. Each of these households contain some form of electricity and/or gas metering data as well as linked demographics, building characteristics and appliance uptake information gathered through household surveying.

The long-term objective of this work is to better understand the complex nature of Australian consumer's behaviour and provide comprehensive information to underpin key decision making, helping to deliver an affordable and sustainable energy future for Australia.

\subsection{Challenges and Related work}
A household’s energy consumption depends on numerous factors such as the socio-demographic characteristics and behaviour of occupants, features of the dwelling, and climate/seasonal factors, among many others \cite{ausResidential,AustralianEnergyUpdate2018,abrahamse2009socio,frederiks2015socio}. In addition,  energy load series is highly non-linear and non-stationary, with several cycles that are nested into each other. All these factors combined together make the task of building accurate prediction models challenging. As such, the underlying function which governs the energy demand is fairly sophisticated, making it difficult to obtain a suitable model by looking only at the input data.

In practice, many time series scenarios are best considered as components of some vector-valued time series having not only serial dependence within each component series, but also interdependence between the other series of elements \cite{brockwell2002introduction}. There are various statistical and machine learning techniques to handle time series predictions. Ensemble learners such as Random Forest and Gradient Boosting are known to be among the most competitive forms of solving time series predictive tasks \cite{oliveira2014ensembles}. The ARIMA (Autoregressive Integrated Moving Average) technique is another popular time series forecasting model which in the past, was mainly used for load forecasting. ARIMA models aim to describe the autocorrelation of the data as well as the error associated with the prediction of the previous time instance \cite{contreras2003arima}.  Neural networks have also received a lot of attention recently due to their ability  to capture the non-linear relationship between the predictor
variables and the target variable \cite{abdel2019accurate}. These methods focus on point forecasting (at time $t$ the task is to predict the load value for  time $t+d$, where d is the forecasting horizon). Predicting an interval of values with a certain probability instead of just predicting a single value gives more information about the variability of the target variable \cite{rana2013prediction}.

GPs are nonparametric Bayesian models which have been employed in a large number of fields for a diverse range of applications \cite{ambrogioni2019complex,pasolli2010gaussian}. The work reported by Swastanto \cite{swastanto2016gaussian} highlights the inflexibility of using parametric models in time series forecasting, thus encouraging the use of nonparametric models. In the nonparametric model, training data is best fitted when constructing the mapping function, whilst maintaining some ability to generalise to unseen data. Nonparametric models use a flexible number of parameters, and the number of parameters are allowed to grow as it learns from more data \cite{hjort2010bayesian}. Therefore, a nonparametric prediction method such as GP is more suitable \cite{rasmussen2003gaussian} for applications such as energy demand forecasting.  Such methods allow the data to speak for itself by requiring minimal assumptions on the data. Further to this, additive models are also popular nonparametric regression methods used for time series forecasting which combine adaptive properties of nonparametric methods and the usability of regression models \cite{thouvenot2015electricity}.

Regression can face issues when there are too few training data points available to learn a good model. Trying to learn a flexible model with many parameters from a few data points may result in over-fitting where the model mistakes artefacts of the specific available samples as actual properties of the underlying distribution \cite{topa2012gaussian,brahim2004gaussian,bonilla2008multi}. We encounter this issue when forecasting energy consumption across Australia, especially when only a few training data points are available for each household. The datasets provide sparse and often irregularly sampled time series instances. Though this is a challenge, there is potential to use the energy consumption data of similar households to help learn the consumption model of each individual household.

There has been a lot of work in recent years on time series prediction with GP \cite{filip2019smooth,mcdowell2018clustering,brahim2004gaussian}. Most of the frameworks developed on GPs have focused on using large datasets \cite{binois2018practical}. In the current study, however, we focus on the other frontier of GP:  when the dataset consists of a considerably large number of short independent instances. The model that we propose takes inspiration from the construction of stacked Gaussian processes \cite{neumann2009stacked,bhatt2017improved} and multi-tasking Gaussian Process learning \cite{bonilla2008multi}, where they take the form of a hierarchical multi-layered Gaussian process, with a stacked kernel.

\subsection{Contribution of our approach}
We construct a simple yet effective Gaussian process framework in the domain of nonparametric supervised learning regression for time series forecasting.
In the proposed stacked GP model, the predictive posteriors of each GP applied on each instance are embedded in the prior and likelihood of the next level stacked GP. Furthermore, kernel functions are designed to cater for the time-varying seasonal patterns, and shared across each instance. Therefore, with our framework, it is possible to extract the underlying structure from the data, even when using short time series data. Another strength of our model is its analytical simplicity, which provides a clear insight into how energy consumption relates to the time series, spatial disparity and other demographic data. Therefore, the model is capable of capturing the seasonal variation of energy consumption along with other related household data. Furthermore, the proposed model has a flexible modelling capability through kernel modification. Hence, it can model a time series with a broad range of complexity.

As our model is based on GP which is a special class of Bayesian probabilistic modelling, it provides a distribution instead of a point forecast. Therefore, we will have a point forecast which can be obtained by the mean of the distribution and its uncertainty (the variance of the distribution). As a result, for each point that is predicted with a GP, we are given the perceived uncertainty of that prediction.

\section{Overview of Gaussian Process}

GP sets an assumption that for every finite set of points $x_{i}$ in $X$, the prior distribution of the vector $(f(x_1),..., f(x_n))$ is multivariate Gaussian. Thus, the response function $f:X\rightarrow Y$ is \lq a-priori\rq \  modelled by the sample path of a GP. Here, we assume some multivariate Gaussian distribution can be used to represent any observation in our dataset. This nonparametric model has a much less restrictive assumption over the function $f$, where we only assume that the function is sufficiently smooth. This smoothness of the function can be defined by tuning the hyper-parameters of a Gaussian process regression. Such a Gaussian distribution can be characterised by defining the mean vector and the covariance function.

We have $n$ observations such that $\mathbf{y}=\{y_1,y_2,...,y_n\}$ which correspond to set of inputs $\mathbf{X} = \{\mathbf{x_1},\mathbf{x_2},...,\mathbf{x_n}\}$ where each $\mathbf{x_i}$ is a $d$-dimensional data point. Assuming that the noise or uncertainty when making an observation can be represented by a Gaussian noise model which has a zero mean and $\sigma_n$ variance, any observation can be written as,
\begin{equation}
y_i = f(\mathbf{x_i}) + \mathcal{N}(0,\sigma_n^2).
\end{equation}

Gaussian distributions are completely parameterized by their mean $m(\mathbf{x})$ and the covariance function $k(\mathbf{x_1}, \mathbf{x_2})$, defined as
\begin{align*}
m(\mathbf{x})=\mathbb{E}[f(\mathbf{x})],\ \ \  k(\mathbf{x_1}, \mathbf{x_2}) = cov \ (f(\mathbf{x_1}), f(\mathbf{x_2}))
\end{align*}
Any collection of function values has a joint Gaussian distribution
\begin{equation}
[f(\mathbf{x_1}),f(\mathbf{x_1})....,f(\mathbf{x_n})]^{\top} \thicksim \mathcal{N}({\mu},{K})
\end{equation}
where the elements in the covariance matrix $\mathbf{K(X,X)}$ are covariance functions between all training inputs, has entries $\mathbf{K(X,X)}_{ij} = k(\mathbf{x}_{i}, \mathbf{x}_{j})$, and the mean ${\mu}$ have entries $\mu_{i} = m(\mathbf{x}_{i})$. The properties of the functions such as smoothness and periodicity are determined by the kernel function, which should be any symmetric and positive semi-definite function \cite{wilson2013gaussian}. In contrast to regression methods where the goal is to find $f(\mathbf{x_i})$, in GP we try to predict $y_i$ whose expectation value is equal to that of $f(\mathbf{x_i})$. Consider that the covariance function (${k}$) is defined using some hyperparameters, $\theta$. Training or learning a GP regression model means determining the optimum set of hyperparameters in the covariance functions.


Standard gradient-based techniques such as conjugate gradient or quasi-Newton methods can be used to determine the best parameters. The prediction for some test inputs $\mathbf{X_*}$ is given by mean $\mathbf{\mu_T}$ and the uncertainty of the prediction is captured by its variance  $\mathbf{\Sigma_T}$ as shown in Eq. (\ref{gp-output}).
\begin{equation}{\label{gp-output}}
\begin{array}{l@{{}={}}l}
\mathbf{\mu_T} & \mathbf{K}(\mathbf{X_{*}},\mathbf{X})[\mathbf{K}(\mathbf{X},\mathbf{X})+\sigma_n^2\mathbf{I}]^{-1}\mathbf{y}\\
\mathbf{\Sigma_T} &
\mathbf{K}(\mathbf{X_{*}},\mathbf{X_*}) - \mathbf{K}(\mathbf{X_{*}},\mathbf{X})[\mathbf{K}(\mathbf{X},\mathbf{X})+\sigma_n^2\mathbf{I}]^{-1}\mathbf{K}(\mathbf{X},\mathbf{X_*})+\sigma_n^2\mathbf{I}
\end{array}
\end{equation}

\section{Proposed data analytic model for energy prediction}

\subsection{Feature selection}
Feature selection is the process of choosing a set of informative variables that are necessary and sufficient for an accurate prediction. Selection of a suitable set of features is one of the critical factors for successful prediction \cite{guyon2003introduction}. Factor analysis can measure the correlation between energy consumption based on the comprehensive data and a large range of factors (including environmental, demographic and other dwelling-specific factors) \cite{jones2015socio}. While a significant amount of literature exists \cite{ausResidential,AustralianEnergyUpdate2018} on the factors affecting household energy use, this step is critical to discern which of these causes would explain the most variance when forecasting. In our case, the features selected are the previous energy  consumption (gas or electricity) of each dwelling, weather information such as air temperature (weather data was collected from Bureau of Meteorology \cite{weather}), HDD (heating degree day), and CDD (cooling degree day) \cite{spinoni2018changes}; and other demographic and dwelling information (income level, number of rooms  etc.). By linking variables such as demographics and weather, we can aim to accurately  forecast future energy use.

Data related to household energy consumption is held by numerous parties. It is also formatted to different standards and access is often restricted. In our previous research, we have conducted single and multi-factor analysis on various such data sets obtained from different energy bodies. However,  in this paper we specifically focus on a dataset obtained from the Australian Energy Regulator's (AER's) 2017 Electricity Bill Benchmark's survey \cite{AER}.

\begin{figure}{}
\centering
\includegraphics[width=12cm]{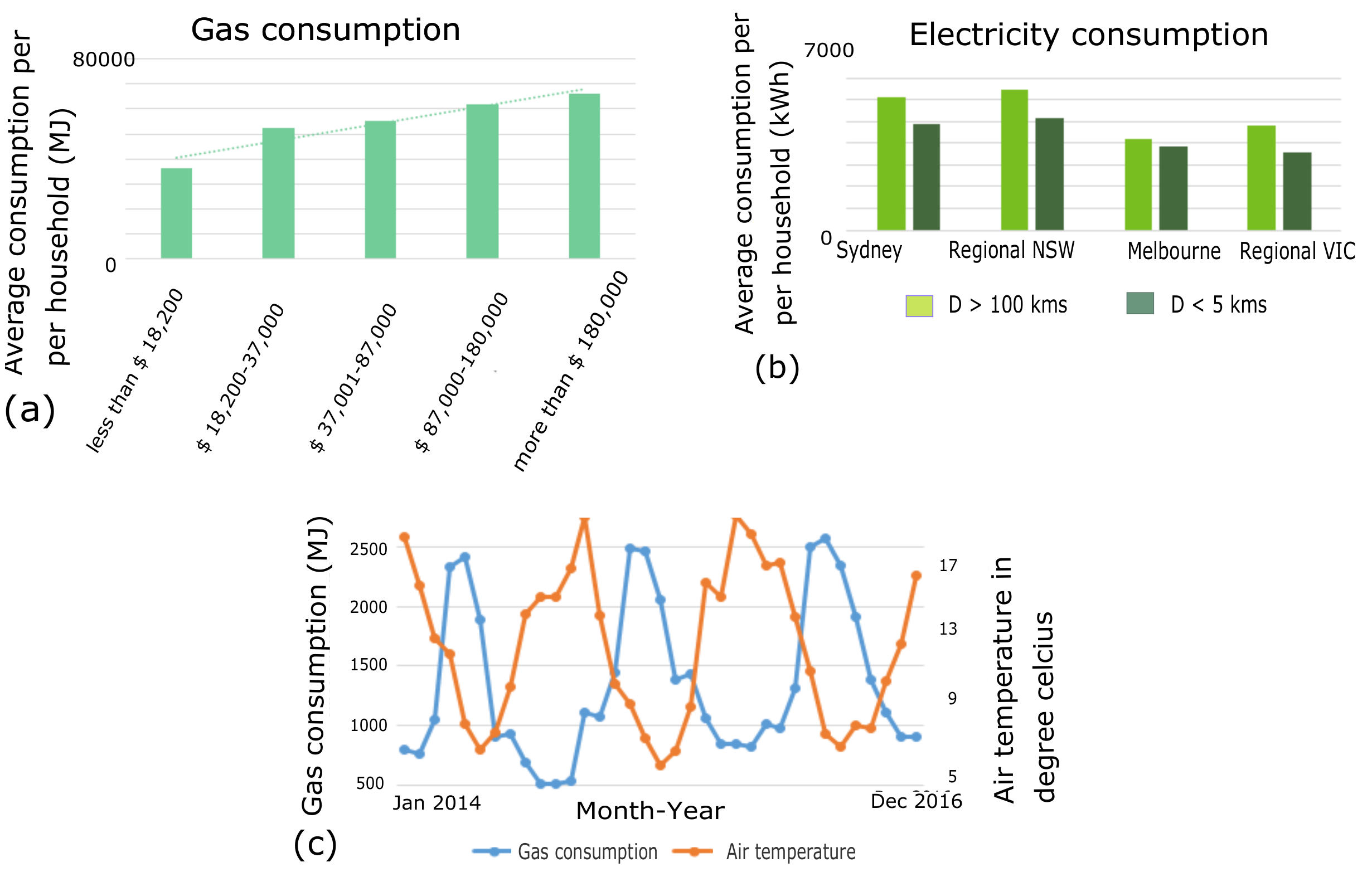}
\caption{\label{demographicData} (a) Average annual gas consumption for different levels of household income, (b) Average annual electricity consumption across regions in NSW and VIC based on distance to the sea, where D denotes distance from the dwelling to the sea in kiolmetres, (c) Relationship between air temperature and household gas consumption over time for a selected dwelling in Australia. In Australia, December-February is summer season and June-August is winter season. }
\end{figure}

We have studied factor analysis extensively on the AER data and other datasets in terms of both gas and electricity consumption. A few examples for demographic and weather related factor analysis outcomes are illustrated in Fig. \ref{demographicData} (a)-(c). Fig. \ref{demographicData} (a) clearly shows that there is a gradual increase in household gas consumption with respect to total household income per annum. Fig. \ref{demographicData} (b) shows the association between household electricity consumption and distance from the sea for dwellings across suburbs in New South Wales (NSW) and Victoria (VIC). It appears that dwellings located further from the sea tend to have higher levels of electricity consumption. It can be observed that the gas consumption is higher during the winter period and gradually decreases with increases in the temperature. This effect is due predominantly due to gas space heating. However, the lower temperatures may also increase the amount of energy consumed by appliances such as gas hot water systems. The relationship between air temperature and gas consumption is illustrated in Fig. \ref{demographicData} (c)for a given dwelling. It can be observed that the gas consumption is higher during the winter season and gradually decreases with the increase of the temperature. This effect is due predominantly due to gas space heating. However, the lower temperatures also increase the amount of energy consumed by appliances such as gas hot water systems.

To this end, we have also identified some of the potential advantages of including these additional factors in our analysis to help explain variation in the energy consumption of different households across various Australian regions. We have the opportunity to explore similarities and differences across regions and measure the effect of various factors on household energy use. This will allow us to develop an improved framework for energy prediction.

\subsection{Modelling for energy forecasting \label{our_method}}
We propose a stacked GP model which combines different models to produce a meta-model with a better predictive performance than the constituent parts. In the context of energy consumption forecasting, our meta-model fuses the seasonally varying pattern of the energy consumption of each household with the other information related to household consumption. Thus, our goal is to fuse data from multiple households with a GP to fully exploit the information contained in the covariates and model spatio-temporal correlations. Fig. \ref{proposed} illustrates the proposed stacked GP model for energy consumption forecasting, where the mean and variance vectors of the predictive posteriors of a set of time series GPs are embedded into the prior and likelihood of the ensemble GP as: $\text{GP}(m(\bf{x}),\textit{k}(\bf{x,x^{\prime}}))$.

\begin{figure}{\label{proposed}}
	\centering
	\includegraphics[width=10cm]{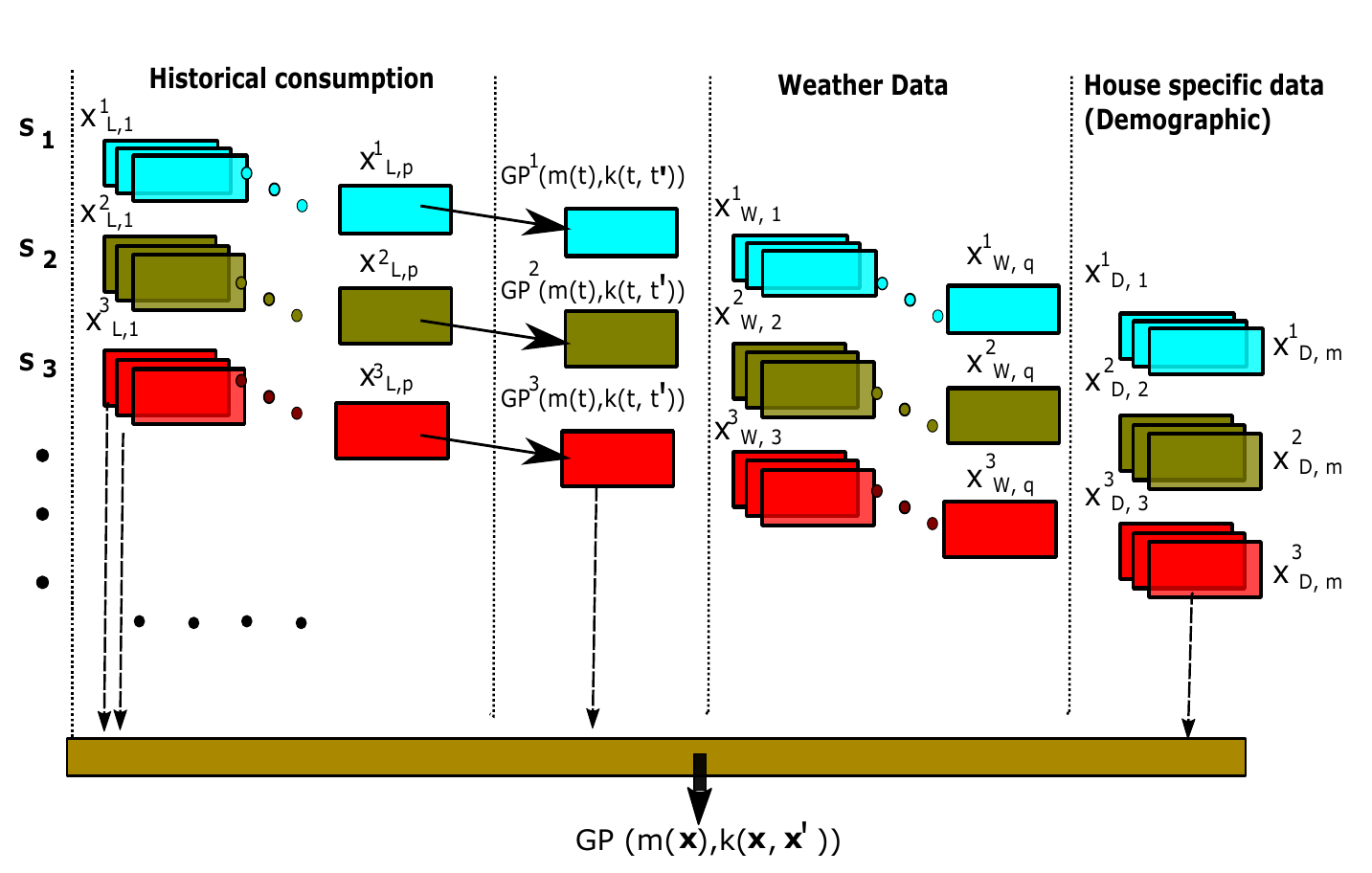}
	\caption{\label{proposed} Schematic of proposed stacked Gaussian process model for energy forecasting. For each household, we take into account previous energy consumption data ($x^i_{L,z}$), weather data ($x^i_{W,z}$) and demographic data ($x^i_{D,z}$) and is differentiated using different colours. The mean and variance vectors computed from the predictive posterior of the time series GP ($x^i_t$)  applied to each instance, ${S_{i}}$ is embedded into the feature vector of the proposed stacked GP model.}
\end{figure}


We will begin by  computing the posterior over all the random functions. As the likelihood and prior are Gaussian, the posterior over functions is also another Gaussian given by:
\begin{eqnarray}{\label{posterior}}
p(\bf{f_t}|\bf{T,y_t})\sim \mathcal{N}(\bf{f_t}|\bf{\bar{\mu}},\bf{\bar{\Sigma}})
\end{eqnarray}
where
\begin{equation}
\begin{array}{l@{{}={}}l}
\mathbf{\bar{\mu}} & \mathbf{K}(\mathbf{T},\mathbf{T})[\mathbf{K}(\mathbf{T},\mathbf{T})+\sigma_n^2\mathbf{I}]^{-1}\mathbf{y_t}\\
\mathbf{\bar{\Sigma}} &
\mathbf{K}(\mathbf{T},\mathbf{T}) - \mathbf{K}(\mathbf{T},\mathbf{T})[\mathbf{K}(\mathbf{T},\mathbf{T})+\sigma_n^2\mathbf{I}]^{-1}\mathbf{K}(\mathbf{T},\mathbf{T})+\sigma_n^2\mathbf{I}
\end{array}
\end{equation}
where $\bf{f_t}$ is a vector containing all the random functions evaluated at training data input vector, $\bf{f_t}=[f(t_{1}),f(t_{2})...,f(t_{n})]^{\top}$. $\bf{T}$ is the time vector and $\bf{y_t}$ denotes the training response variable (in our case energy consumption). To add some flexibility to capture time varying seasonality,  a local kernel (squared-exponential kernel) is combined with a periodic kernel. This will allow us to model functions that are only locally periodic and change over time. $k(t,\bar{t})$ is given by
\begin{eqnarray}{\label{kernels}}
k({t},{\bar{t}})=\sigma^{2}\exp\Big(\frac{-2 \sin^{2}(\pi|t-\bar{t}|/p\Big)}{l^2})+\exp\Big(\frac{-(t-\bar{t})^2}{2 l^2}\Big)
\end{eqnarray}
where $\sigma,l,p$ denote hyper-parameters, $t$ represents the time, and $*$ denotes the testing data. We can now predict a household's energy consumption at any given point in time by computing the predictive posterior:
\begin{eqnarray}{\label{predictive_posterior}}
p(\bf{y_{*,t}|\bf{T_{*},T,y_t}}) = \int \textit{p} (\bf{y_{*,t}|\bf{T_{*},f_{t},T}})\textit{p} (\bf{f_{t}|\bf{T,y_t}}) d\bf{f_{t}}
\end{eqnarray}
\begin{eqnarray}{\label{output}}
p(\bf{y_{*,t}|\bf{T_{*},T,y_t}})\sim \mathcal{N} (\bf{y_{*,t}|\bf{\bar{\mu}_{*,t}},\bar{\Sigma}_{*,t}})
\end{eqnarray}

This is applied for each task (denoted by $i$, where task is a sample time series drawn from one dwelling) and the mean and variance are calculated from the below formula:
\begin{eqnarray}{\label{mean}}
\bar{\mu}_{*,t}^i=\mathbf{K}^{i}(\mathbf{T_*},\mathbf{T})[\mathbf{K}^{i}(\mathbf{T},\mathbf{T})]^{-1}\mathbf{y_t}^{i}
\end{eqnarray}
\begin{eqnarray}{\label{variance}}
\bar{\Sigma}_{*,t}=\mathbf{K}(\mathbf{T_{*}},\mathbf{T_{*}}) -\mathbf{K}(\mathbf{T_{*}},\mathbf{T})[\mathbf{K}(\mathbf{T},\mathbf{T})+\sigma_n^2\mathbf{I}]^{-1}\mathbf{K}(\mathbf{T},\mathbf{T_{*}})+\sigma_n^2\mathbf{I}
\end{eqnarray}

The mean  $\bar{\mu}_{*,t}^i$ and variance $\bar{\Sigma}_{*,t}^i$ obtained from the predictive posterior of each task are taken into the next stage,\\
$(\mathbf{\bar{\mu}_{*,t}},\mathbf{\bar{\Sigma}_{*,t}})=\\
\{(\bar{\mu}_{*,t}^1,\bar{\Sigma}_{*,t}^1)..,(\bar{\mu}_{*,t}^i,\bar{\Sigma}_{*,t}^i)..,(\bar{\mu}_{*,t}^n,\bar{\Sigma}_{*,t}^n)\}=
\{(\mathbf{\bar{\mu}_{*,t}^{Tr}},\mathbf{\bar{\Sigma}_{*,t}^{Tr}}),(\mathbf{\bar{\mu}_{*,t}^{Te}},\mathbf{\bar{\Sigma}_{*,t}^{Te}})\}$
\begin{eqnarray}{\label{output-gp1}}
\mathbf{X_{t}^{Tr}}=(\mathbf{\bar{\mu}_{*,t}^{Tr}},\mathbf{\bar{\Sigma}_{*,t}^{Tr}})\ \ \text{if}\ \ \text{MPE} <\tau, \nonumber \\
\mathbf{X_{t}^{Te}}=(\mathbf{\bar{\mu}_{*,t}^{Te}},\mathbf{\bar{\Sigma}_{*,t}^{Te}})
\end{eqnarray}
$Tr$ and $Te$ represent training and testing data. Only the tasks with $\text{MPE}$ (Mean Percentage Error) $<\tau$ are considered for model training. $\tau$ can be learnt from the training data. $\mathbf{X_{t}^{Tr}},\mathbf{X_{t}^{Te}}$ are then used in the prior and likelihood of the  predictive posterior in the next stage.

At the next stage of our proposed stacked model, let's state that the function $f_{st}$ is random and drawn from a multivariate Gaussian distribution, where $i$ and $j$ represent different tasks (or dwellings). As such:

\begin{equation}
\begin{bmatrix} f_{st}(\mathbf{x}_{i}) \\ f_{st}(\mathbf{x}_{j}) \\ \vdots \end{bmatrix}
\sim
\mathcal{N} \left(\begin{bmatrix} m_{st}(\mathbf{x}_{i}) \\ m_{st}(\mathbf{x}_{j}) \\ \vdots \end{bmatrix},
\begin{bmatrix} k_{st}(\mathbf{x}_{i},\mathbf{x}_{i}) & k_{st}(\mathbf{x}_{i},\mathbf{x}_{j}) & \cdots\\ k_{st}(\mathbf{x}_{j},\mathbf{x}_{i}) & k_{st}(\mathbf{x}_{j},\mathbf{x}_{j})\ \ & \cdots \\ \vdots & \vdots & \ddots \end{bmatrix}\right)
\end{equation}

where $m$ and $K$ are the mean and covariance function, respectively. Since the function $f_{st}$ is random,
the optimum $f_{st}$ will be inferred via the Bayesian inference. Therefore, we can compute the predictive posterior over all the random functions considering all the tasks.
\begin{eqnarray}{\label{posterior-gp2}}
p(\bf{y_{*,st}|\bf{X_{*},X,y}})=\int\textit{p} (\bf{y_{*,st}|\bf{X_{*},f_{st},X}})\textit{p} (\bf{f_{st}|\bf{X,y}})d\bf{f_{st}}
\end{eqnarray}
here $\bf{y_{*,st}}$ is the predictive posterior of the proposed stacked GP. \\ $\mathbf{X}=\{\mathbf{X_{L}^{Tr}},\mathbf{X_{W}^{Tr}},\mathbf{X_{D}^{Tr}},\mathbf{X_{LT}^{Tr}},\mathbf{X_{t}^{Tr}}\}$, representing training data for historical load ($\mathbf{X_{L}^{Tr}}$), weather ($\mathbf{X_{W}^{Tr}}$), demographic ($\mathbf{X_{D}^{Tr}}$), lag time ($\mathbf{X_{LT}^{Tr}}$) and mean predictive posterior of time series GP ($\mathbf{X_{t}^{Tr}}$) respectively. $\bf{X_{*}}$ denotes the testing data. The predictive distribution is again Gaussian, with a mean $\bf{\mu_{*,st}}$, and covariance $\bf{\Sigma_{*,st}}$.
\begin{eqnarray}{\label{output}}
p(\bf{y_{*,st}|\bf{X_{*},X,y}})\sim \mathcal{N} (\bf{y_{*,st}|\bf{\mu_{*,st},\Sigma_{*,st}}})
\end{eqnarray}
where, $\bf{\mu_{*,st},\Sigma_{*,st}}$ can be obtained from Eq.(\ref{gp-output}).

The following proposed kernel is used to model the inter-sample relationship.
\begin{eqnarray}{\label{covariance}}
k_{st}(i,j)=\prod_{z=1}^{z=p} k(x_{\text{L},z}^i,x_{\text{L},z}^j)\times \prod_{z=1}^{z=q} k(x_{\text{W},z}^i,x_{\text{W},z}^j)\\ \nonumber
\times \prod_{z=1}^{z=m} k(x_{\text{D},z}^i,x_{\text{D},z}^j)\times k(x_{\text{LT}}^i,x_{\text{LT}}^j)\times k(x_{\text{t}}^i,x_{\text{t}}^j)
\end{eqnarray}

Here, $k(x_{\text{L},z}^i,x_{\text{L},z}^j), k(x_{\text{W},z}^i,x_{\text{W},z}^j), k(x_{\text{D},z}^i,x_{\text{D},z}^j), k(x_{\text{LT}}^i,x_{\text{LT}}^j)$, and  $k(x_{\text{t}}^i,x_{\text{t}}^j)$ denote covariates generated by historical consumption load, weather data, demographic data, time stamp of the lag time and the mean values obtained from the time series GP respectively. $p$ is the number of months, $q$ is the total number of weather data points (e.g. $x_{\text{L},z}^i$ denotes the energy consumption of household $i$ in the $z^{th}$ month). These notations are illustrated in Fig. \ref{proposed}.  Intuitively, if two tasks, $\mathbf{x}_{i}$ and $\mathbf{x}_{j}$ are similar, then $f_{st}(\mathbf{x}_{i})$ and $f_{st}(\mathbf{x}_{j})$ should also be similar, which explains why the function generated by a GP is smooth. Our assumption of this similarity or smoothness is encoded by the kernel function $k(\mathbf{x}_i,\mathbf{x}_j)$.

\section{Experimental setup}
\subsection{Data Description and Preparation}
In our study, we have used a dataset obtained from the AER to demonstrate how our stacked GP model learns spatio-temporal patterns in the data.  We investigate two cases: gas consumption and electricity consumption forecasting. Our primary focus is on gas use forecasting where we have short  but multiple time series data (gas data is collected from basic meters, thus the dataset is sparse. To overcome this issue, we have drawn multiple short time series data for our experiments). Nevertheless, in order to further demonstrate the capability of our model, we have also applied the proposed stacked GP model on household electricity data.

\begin{enumerate}
  \item \textbf{Case 1:} The AER gas consumption dataset consists of quarterly gas consumption data for approximately 2600 dwellings, with data spanning from year 2013-2016. Though this dataset consists of quarterly consumption readings, the duration of a quarter is not consistent across years nor within individual household. Thus, a measure of monthly gas usage was generated based on daily average usage. Thus, monthly gas usage was generated based on daily average usage. Table \ref{tab:DataGasElec} shows the statistics for our experimental set-up, including the mean gas consumption and number of samples in the training and testing data split across major states (e.g. NSW, VIC, QLD, ACT)  and focusing only on the state of Victoria (VIC). Our reason for examining Victoria separately from other states (i.e. \lq Overall\rq \ in Table \ref{tab:DataGasElec})   is due to AEMO’s current interest in understanding  VIC's energy consumption as a stepping stone to understanding Australia’s consumption as a whole. This is in part because smart meter data is in such high abundance in Victoria due to the state government’s smart meter roll-out \cite{energyvic} and in part due to the high penetration of gas appliances in Victoria. We then split the overall population of the data into training and testing based on the year. The years 2013-2015 are used for training and the year 2016 is used for testing.
  \item \textbf{Case 2:} AER smart meter electricity data consists of about 3100 households situated in VIC and NSW. The data spans from year 2015-2016. We generated monthly electricity consumption for each household based on half-hourly consumption data. Table \ref{tab:DataGasElec} provides detailed statistics of the experimental set-up.

\end{enumerate}









\paragraph{Evaluation metrics}To measure predictive accuracy, we use two standard performance measures: Mean Absolute Error (MAE) and R-squared ($\text{R}^{2}$).
\begin{equation}{\label{MAE}}
\text{MAE}=\frac{1}{n}\sum_{i=1}^{n}|y_{i}-{f_{i}}|
\end{equation}

\begin{equation}{\label{Rsquared}}
\text{R}^2=1-\frac{\sum_{i=1}^{n}(y_{i}-{f_{i}})^2}{\sum_{i=1}^{n}(y_{i}-\bar{y})^2}
\end{equation}
where $y_{i}$ and $f_{i}$ are the actual and predicted gas or electricity consumption at time $i$, and $n$ is the total number of predicted loads.



\subsection{Baselines and other machine learning models used for comparison\label{baseline}}
We compare our proposed stacked GP method against the following commonly used machine learning techniques. We use the same feature set for all the methods.
\begin{enumerate}
  \item Random Forest (RF): RF is a supervised learning algorithm which builds multiple decision trees to obtain more accurate and stable results \cite{Breiman2001}.
 \item Gradient Boosting (GB): GB uses a boosting technique which sequentially ensembles trees into a single strong learner \cite{friedman2002stochastic}.
 \item Multivariate Long Short-Term Memory (LSTM):  This is a recurrent neural network (RNN) architecture which consists of feedback connections and widely used in time series forecasting \cite{li2019ea}.
  \item Auto-regressive (AR): AR models learn from a series of timed steps and takes measurements from previous actions \cite{akaike1969fitting}.
  \item Auto-regressive integrated moving average (ARIMA): ARIMA uses the regression error as a linear combination of error terms whose values occurred contemporaneously and at various times in the past \cite{das1994time}.
  \item Traditional time-series GP: Application of GP to a time series sequence, where predictor variable is a sequence of time instances and response variable is energy consumption (in our case). This method is also described from Eq. \ref{posterior}- \ref{predictive_posterior}, in Sec. \ref{our_method}.

\end{enumerate}








\begin{table}
\caption{\label{tab:DataGasElec} Data statistics for gas and electricity consumption.}
\vspace{0.2cm}
	\begin{center}
		\begin{tabular}{|P{25mm}|P{20mm}|P{20mm}|P{20mm}|P{20mm}|}
			\hline
			\multirow{3}{*}{Training/Testing} & \multicolumn{2}{c|}{Gas} & %
			\multicolumn{2}{c|}{Electricity} \\
			\cline{2-5}
			& \multicolumn{2}{c|}{Region} & \multicolumn{2}{c|}{Region}  \\
			\cline{2-5}
			& VIC & Overall & VIC & NSW   \\
			\hline
			 Training load mean& 4100 MJ & 3600 MJ&  576 kWh & 560 kWh \\
			\hline
			 Testing load mean& 4500 MJ& 3700 MJ& 409 kWh& 527 kWh \\
			\hline
			Training samples&  4586 & 7481 &  6031 & 3499 \\
			\hline
			Testing samples&8882 & 15774 &5650& 2191 \\
			\hline
		\end{tabular}
	\end{center}
	
\end{table}

\section{Results and discussion}
In energy forecasting, we are primarily interested in the predictive accuracy of the model rather than the parameter of the function. Therefore, instead of inferring the parameters to get the latent function of interests (as in parametric methods), here we  infer the function $f_{st}$ given in Eq.(\ref{posterior-gp2}) directly, as if the function $f_{st}$ is the \lq parameters\rq \ of our model. We have inferred the function $f_{st}$ from the data through the Bayesian inference to obtain  a probability distribution.

We performed the time series GP prediction by applying a single GP on each data sample, as given in Eq. (\ref{posterior}-\ref{mean}). In order to capture the periodic random functions that vary over time, we used an exponential kernel with a periodic kernel as given in Eq. (\ref{kernels}). Training samples of MPE less than $\tau=1$ are taken to the next stage and embedded in the prior and likelihood in the proposed stacked GP, as given in Eq. (\ref{posterior-gp2}).

\begin{figure}{\label{gas_mae_r2}}
\centering
\includegraphics[width=12cm]{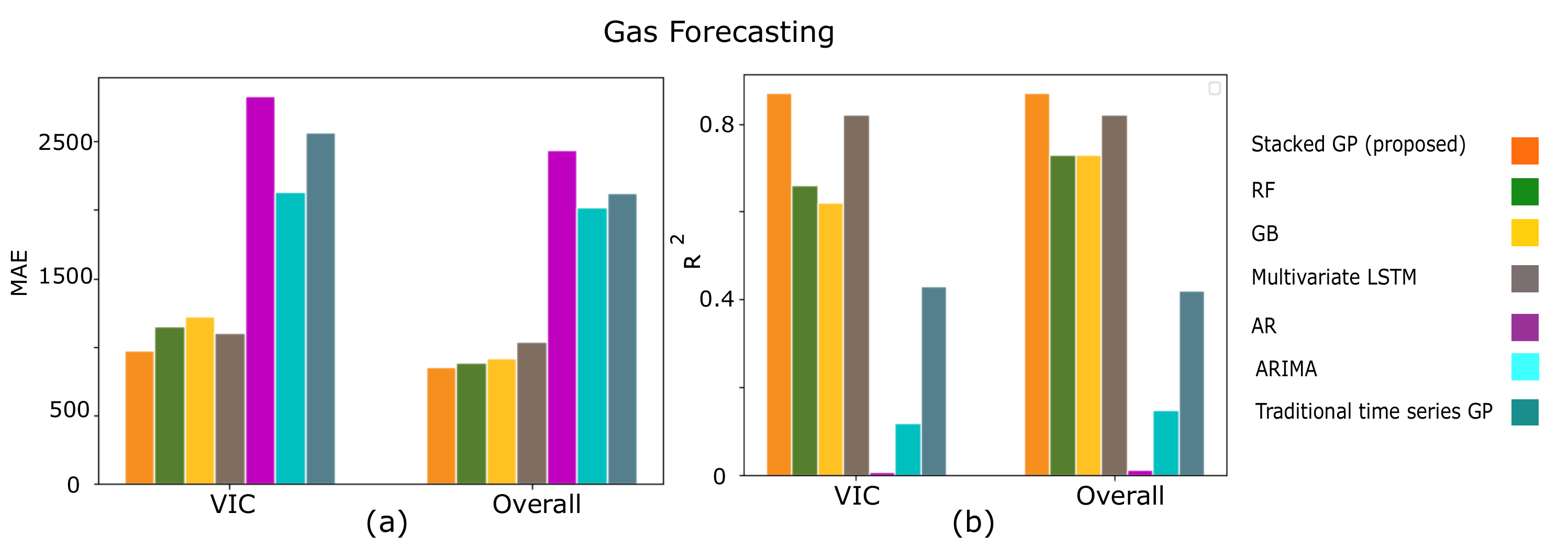}
\caption{\label{gas_mae_r2}  Results: (a) MAE (Mean Absolute Percentage Error) and (b) $R^2$ values for gas consumption forecasting, first for the state of VIC and then overall across Australia. For both MAE and  $R^2$ metrics, the proposed Stacked GP method produced the best results (i.e. smaller MAE and higher  $R^2$) when assessing both Victoria only and Overall.}
\end{figure}


\begin{figure}{\label{elec_mae_r2}}
\centering
\includegraphics[width=12cm]{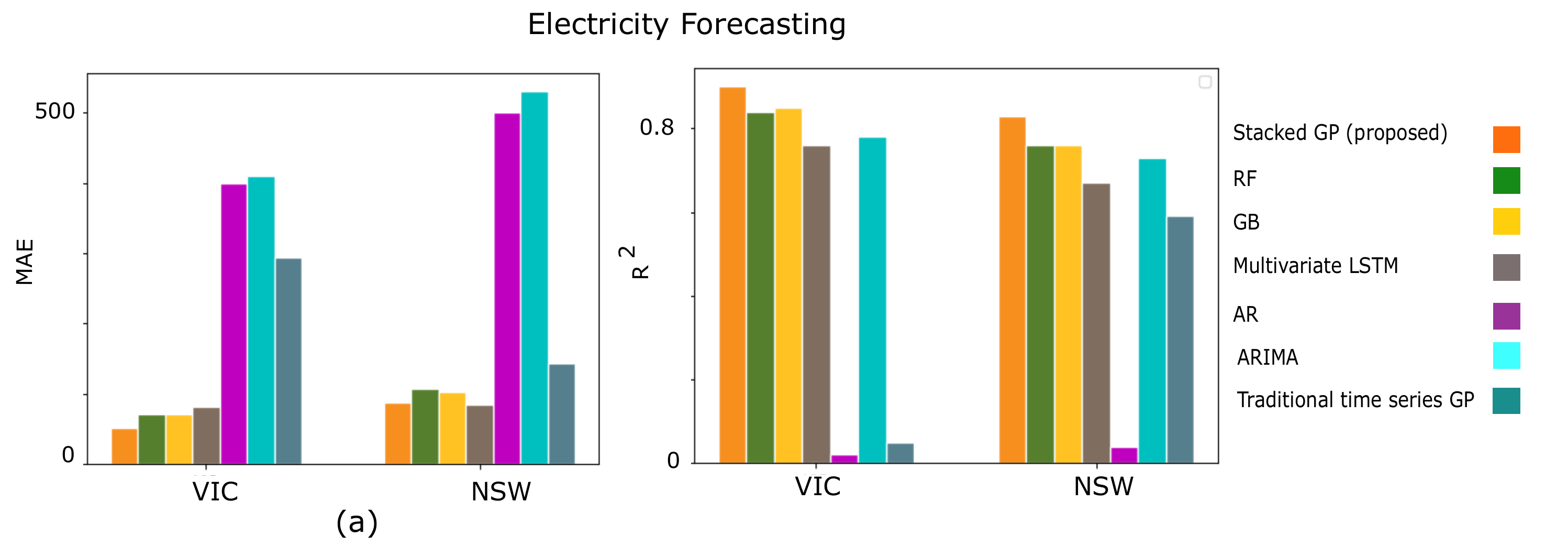}
\caption{\label{elec_mae_r2} Results: (a) MAE and (b) $R^2$ values for VIC and NSW electricity consumption forecasting. For both MAE and $R^2$ metrics, the proposed Stacked GP method produced the best results (i.e. smaller MAE and higher $R^2$) when assessing both VIC and NSW}
\end{figure}



To evaluate the effectiveness of the proposed method, we compared its predictive performance with several machine learning based and naive techniques (Sec. \ref{baseline}).
Grid search is used for hyper-parameter tuning for all the machine learning techniques. The lowest MAE is reported for the proposed stacked GP method (see Fig. \ref{gas_mae_r2} (a) and Fig. \ref{elec_mae_r2} (a)). In terms of the $R^2$ metric, the proposed method outperforms all the other techniques as shown in the Fig. \ref{gas_mae_r2} (a) and Fig. \ref{elec_mae_r2} (b).

\begin{figure}{}
\centering
\includegraphics[width=\textwidth]{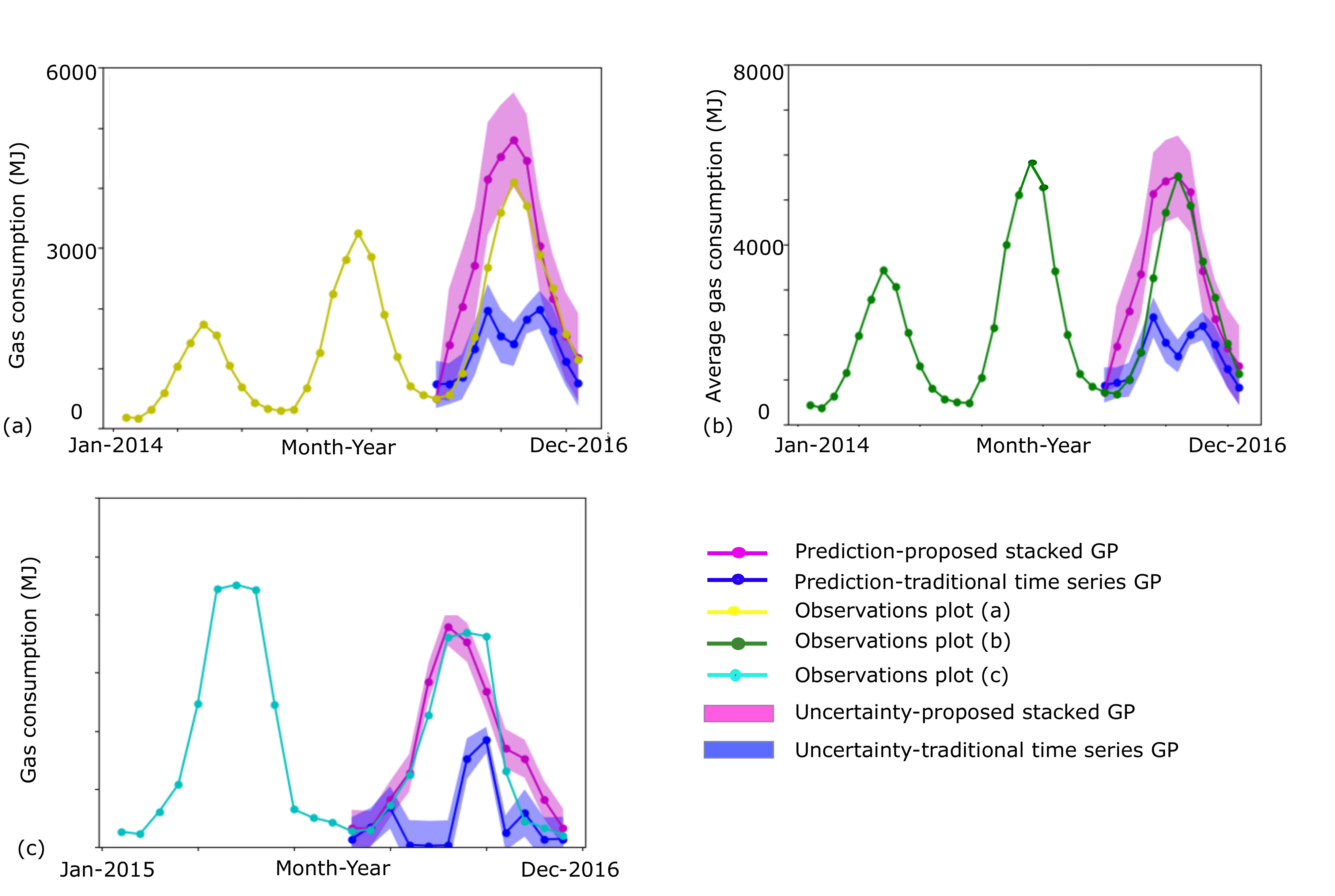}
\caption{\label{gas_prediction_actual}Results comparison of the proposed method against the traditional time series GP. (a) Average overall gas consumption per dwelling in Australia, (b) Average overall gas consumption per dwelling in VIC, Australia. 2014-2015 data used for training and 2016 data used for model evaluation (in (a) and (b) monthly average of the mean and variance of proposed stacked GP output of dwellings across Australia and dwellings located only in VIC are considered), (c) Gas consumption model testing for a median performing dwelling.}
\end{figure}

Further to this, for illustrative purposes, we have depicted the average mean and variance of the proposed model output from all the dwellings across Australia (Fig. \ref{gas_prediction_actual} (a)) and those located in VIC ( Fig. \ref{gas_prediction_actual} (b)), for gas consumption. These figures demonstrate the ability of the proposed model to capture the trend in household consumption patterns. The latest report from AEMO \cite{aemoreport18}, states that the accuracy of AEMO's 2017-18 annual operational consumption forecast is approximately $R^2=0.80$. Though this $R^2$ value is lower than the one reported in this paper, AEMO's value provides forecasting accuracies for state-level energy consumption. This is in contrast to our research based on household-level energy forecasting. As such these may not be directly comparable.

It can also be observed that popular time series forecasting methods such as AR and ARIMA do not perform well in our case, as we have only limited amount of time series data for each task. The shortcomings of these two techniques are caused by their limited assumptions over the underlying function. The ARIMA model assumes that $y_{t+1}$ is linear in the past data and past errors. These linear assumptions are inadequate to model a complex forecasting problem and thus a more powerful model is needed.

It is also important to note that the application of time series GP on individual households provides weak performance metrics due to the limited number of training data samples. This is illustrated in Fig. \ref{gas_prediction_actual} (c) with the aid of a median performing dwelling, where the mean prediction and the variance of the prediction from the proposed stacked GP are compared against the output from the traditional time-series GP. The main motivation of this work is the handling of multiple short time series tasks which require several decisions in terms of how we describe the dynamics of the observed values of other related tasks. Settling on a single answer to these decisions or assumptions over the underlying functions  may be risky in real world time series where one frequently observes changes in properties such as demographics or the weather. The proposed stacked GP method which uses mean and variance from time series GP and considers the inter-household relationship shows superior performance compared to the other techniques considered.





\section{Conclusion}
In this paper, we proposed a nonparametric Bayesian model which uses Gaussian Process (GP) with an ensemble learning approach to forecast energy consumption at a household level across major states in Australia. Our proposed stacked GP model inherits the advantages from both time series GP and multi-dimensional GP, without any prior knowledge about the underlying functional interplay among the latent temporal dynamics. This model was formulated based on the GP predictive distribution over the secondary tasks given the primary task. Here, the primary tasks are formulated as a set of time series functions and the secondary task models the inter-sample relationship using task related descriptors. Experiments on real-world datasets demonstrate that the proposed model is effective and robust to model spatio-temporal patterns in time series data and performs consistently better compared to other machine learning and time series forecasting techniques.

This is one of our  ongoing predictive analytic projects in the energy domain, which is being carried-out in close collaboration with two leading government bodies in Australia. This highlights Australia's efforts in using  machine learning to provide meaningful information that will unlock the mysteries of Australia's energy behaviour – and help to deliver an efficient energy future.

\section{Acknowledgement}
We sincerely thank AEMO for providing valuable feedback on this research and AER for providing access to the valuable dataset used herein. We would also like to thank Dr. Elisha Frederiks (Energy-CSIRO),  Peter Goldthorpe (Energy-CSIRO) and  Dr. Nariman Mahdavi (Energy-CSIRO) for their constructive and insightful feedback to the manuscript.

%
%
%
\bibliographystyle{splncs04}
\bibliography{references}

\end{document}